\documentclass{article}
\usepackage{spconf,amsmath,epsfig}
\usepackage{graphicx}
\usepackage{subcaption}

\usepackage[compact]{titlesec}
\title{Primitive-based 3D Building Modeling, Sensor Simulation, and Estimation}
\name{Xia Li* \thanks{*The first three authors contributed equally.}, Yen-Liang Lin*, James Miller*, Alex Cheon, Walt Dixon}
\address{GE Global Research, Niskayuna, USA}
\begin{document}
%\ninept
%
\maketitle
\begin{abstract}

As we begin to consider modeling large, realistic 3D building scenes, it becomes necessary to consider a more compact representation over the polygonal mesh model.
%In this work, we explore to transform this dense 3D representation into a compact collection of primitive modeling elements. 
%
Due to the large amounts of annotated training data, which is costly to obtain, we leverage synthetic data to train our system for the satellite image domain.
%
%To simulate varied building shapes and types, we iteratively divide the simulation space into regions, and synthesize 
%different heights and primitive types for each region.  
%
By utilizing the synthetic data, we formulate the building decomposition as an application of instance segmentation and primitive fitting to decompose a building into a set of primitive shapes. 
Experimental results on WorldView-3 satellite image dataset demonstrate the effectiveness of our 3D building modeling approach.
%
% We beat the phase 1a performance goals, and were one of 2 teams awarded the next phase of the program.
% Our method is one of the top two performing approaches in phase 1.a IARPA CORE3D challenge. 
%
\end{abstract}
%\begin{keywords}
%\end{keywords}

\section{Introduction}
\label{sec:intro}
Reconstructing realistic 3D building models from remote sensor data benefits to several tasks including physical security vulnerability assessment, mission planning, and urban visualization, etc.
A primitive based representation provides several advantages over the polygonal mesh representation, such as regularization through prior knowledge, compact representation, and symbolic representation. 
However, building modeling and primitive fitting are still challenging tasks where some questions needed to be addressed, e.g., how many primitives are needed to represent the structure, how those primitives are arranged, and how to determine the best fitting.

%%%%%%%%%%%%%%%%%%%%%%%%%%%%%%%%
% Figure: figure1 
%%%%%%%%%%%%%%%%%%%%%%%%%%%%%%%%
\begin{figure}[!t]
	\includegraphics[width=0.5\textwidth]{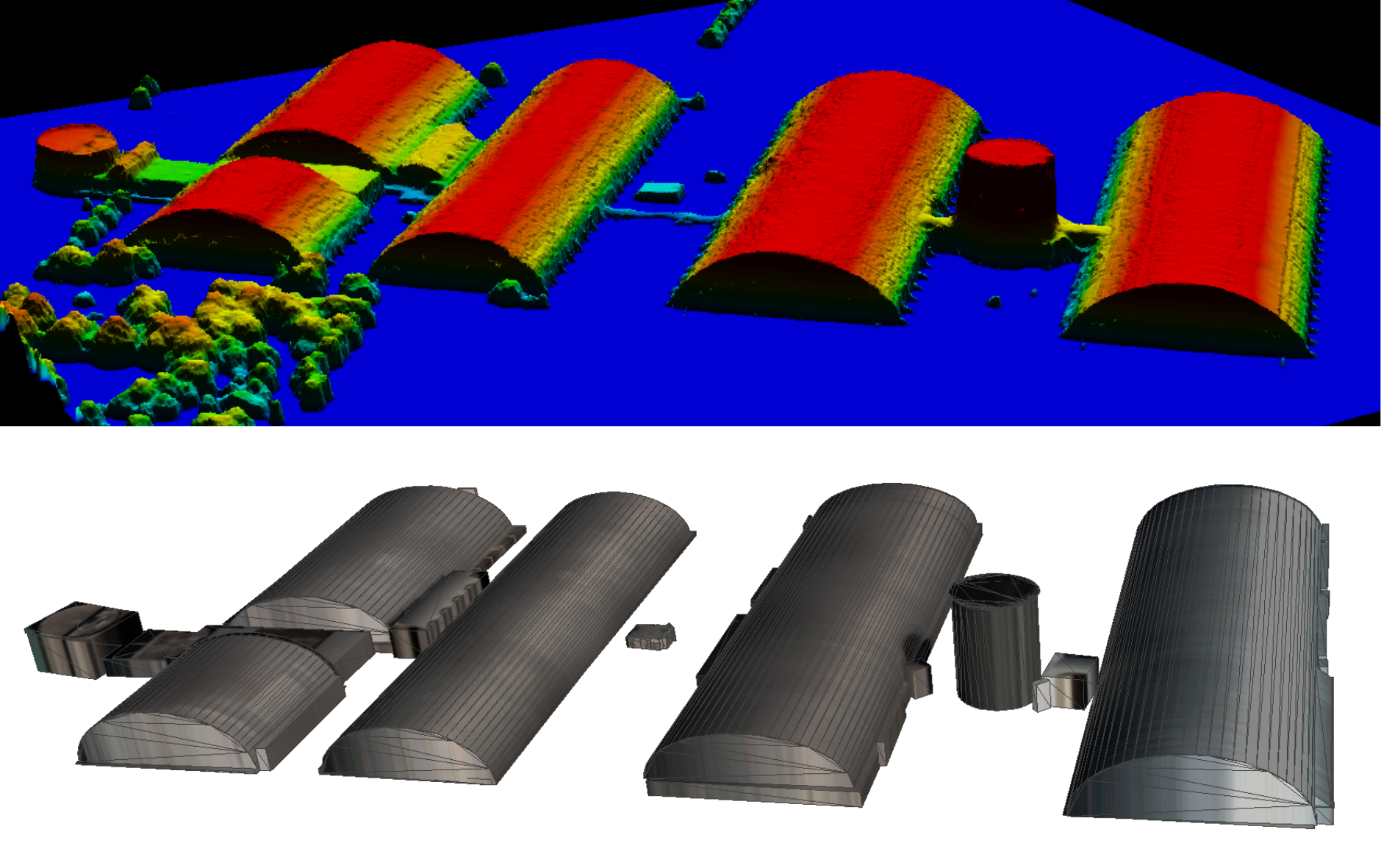}
	\caption{The goal for our system is to represent buildings and other man-made structures from the reconstructed digital height models (DHM) with a collection of geometric primitives.}
	\label{fig:figure1}
\end{figure}
 
An existing work \cite{schnabel2007efficient} utilizes a random sample consensus (RANSAC) to estimate the planes for building walls.
However, RANSAC involves needing to solve many constraints and can run into instability when these constraints contain some amount of noise.
Convex decomposition \cite{ren2011minimum} is another possible approach for shape composition. Ren et al. \cite{ren2011minimum} decomposes arbitrary 2D and 3D shapes into a minimum number of near-convex parts. However, the decomposition is not guaranteed to be formed by primitive shapes.  
One recent approach from \cite{TulsianiCVPR2017} learns to assemble objects using volumetric primitives. 
The parameters of primitives (cuboids), such as the numbers, size and orientation, are estimated via a deep learning network and the obtained reconstruction allows an interpretable representation for the input object.
However, it is an unsupervised approach that requires large-scale training images for each category and cannot accurately fit into the input 3D data. 
Moreover, their method only applies on cuboid representation, limits it ability to more complex building shapes. 

We propose to synthesize training data for primitive-based 3D building modeling, which does not incur costly annotations and allows deep learning models to learn the shape decomposition in a data-driven manner. 
In particular, we present a synthesis pipeline to generate varied building shapes and types.
By utilizing the synthetic data, we formulate the building decomposition as an application of instance segmentation and primitive fitting to decompose a building into a set of sections. 
Each section is then classified as a certain primitive type, and model fitting is applied for adjusting pose and scale of the predicted section.  
%
%%%%%%%%%%%%%%%%%%%%%%%%%%%%%%%
% Figure: system pipeline
%%%%%%%%%%%%%%%%%%%%%%%%%%%%%%%%
\begin{figure*}[t]
	\includegraphics[width=\textwidth]{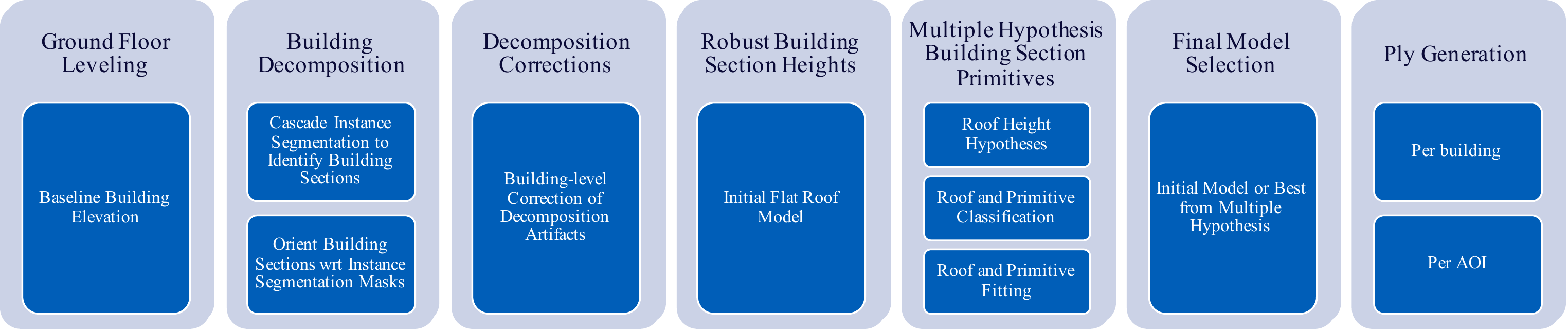}
	\caption{System pipeline of our approach.}
	\label{fig:pipeline}
\end{figure*}

The main contributions of this work include:
\begin{itemize}
	\item We show that  leveraging synthetic data is an effective approach for building decomposition and primitive fitting. In particular, we achieve 
		 promising performance on WorldView-3 satellite image dataset.	
		   
	\item We propose a synthesis pipeline that generates building shapes and types in an iterative manner, which partitions the simulation region into randomly sized non-overlapping regions and synthesize different heights and primitive types for each region.

	\item We formulate the problem of primitive-based 3D building modeling as an application of instance segmentation and primitive fitting to decompose a building into a set of primitive shapes. 
\end{itemize}

\section{Proposed method}
The system pipeline of building 3D primitive models is shown in Figure ~\ref{fig:pipeline}. 
A baseline building elevation (digital terrain model) is estimated and the building model is referenced to this level datum. 
Then we formulate the decomposition problem as an application of cascading instance segmentation, which is extended to decompose a building into a set of sections. 
To improve the decomposition, a correction approach is used to fill gaps interior to a building between individual sections. 
For each section, primitive classification and fitting are applied based on multiple building height hypothesis. 
The best fitting model is selected and 3D model is generated.

\subsection{Building Simulation}
\label{sec:building_simulation}
%%%%%%%%%%%%%%%%%%%%%%%%%%%%%%%%
% Figure: building generation
%%%%%%%%%%%%%%%%%%%%%%%%%%%%%%%%
\begin{figure}[h]
	\includegraphics[width=0.5\textwidth]{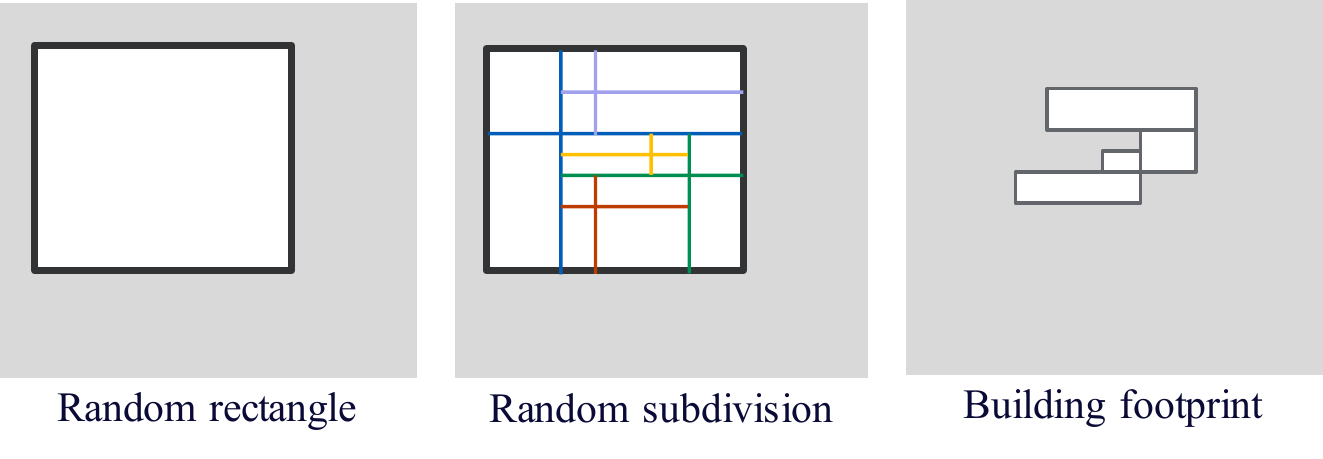}
	\caption{Building section generation.}
	\label{fig:building_generation}
\end{figure}

%%%%%%%%%%%%%%%%%%%%%%%%%%%%%%%%
% Figure: building simulation
%%%%%%%%%%%%%%%%%%%%%%%%%%%%%%%%
\begin{figure}[h]
	\includegraphics[width=0.5\textwidth]{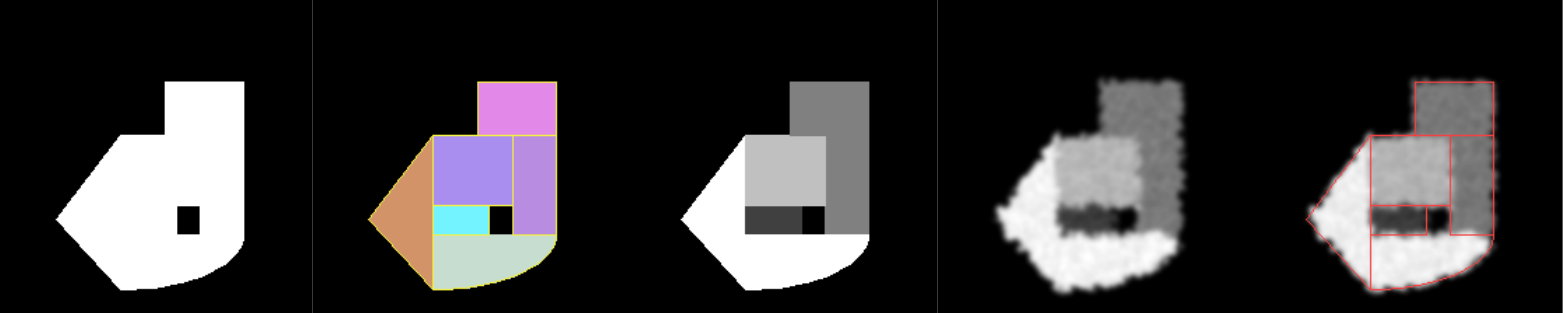}
	\caption{Simulating buildings with multiple primitive types (rectangular prisms, triangular prisms, elliptical cylinders), heights and boundary distortions. Left to right: binary mask footprint, reference primitives, flat roof height map simulation, boundary distortion modeling stereo reconstructions, same with primitives overlaid.}
	\label{fig:building_simulation}
\end{figure}

To simulate a building shape, we define a region of space and recursively randomly partition the region. 
In a manner similar to constructing a quadtree, we randomly sample a point within the region and divide the region into 4 rectangular regions.  
We iterate to partition the region into randomly sized non-overlapping rectangles. 
We then randomly select a subset of the rectangles to form the building, and discard other rectangles. 
The building rectangles, while forming a realistic footprint for a building, typically have more primitives than necessary to represent the building. 
We simplify the selected collection of rectangles by merging adjacent rectangles that completely share an edge (cf. Figure \ref{fig:building_generation}). We assign random heights to each building section and assign roof models to each section.
%
%Furthermore, we simulate the appearance of building footprints using an image-based approach in stereo reconstructed images. 
\subsection{Stereo Simulation}
We simulate how a building will appear in a stereo reconstruction to produce a new height map (cf. Figure \ref{fig:building_simulation}).  First, we generate an ideal height map for the building. Then, we add random Gaussian noise to the simulated heights.  We perturb the boundaries of the building and the boundaries between building sections by randomly dilating points along the height map to model the boundary properties observed in stereo reconstructions from tools like s2p. Finally, we smooth the noisy and perturbed height map to model the correlation typically seen in the output of satellite stereo reconstruction tools like s2p. 

\subsection{Building Decomposition}
%%%%%%%%%%%%%%%%%%%%%%%%%%%
% Figure building decomposition
%%%%%%%%%%%%%%%%%%%%%%%%%%%
\begin{figure}[h]
	\includegraphics[width=0.5\textwidth]{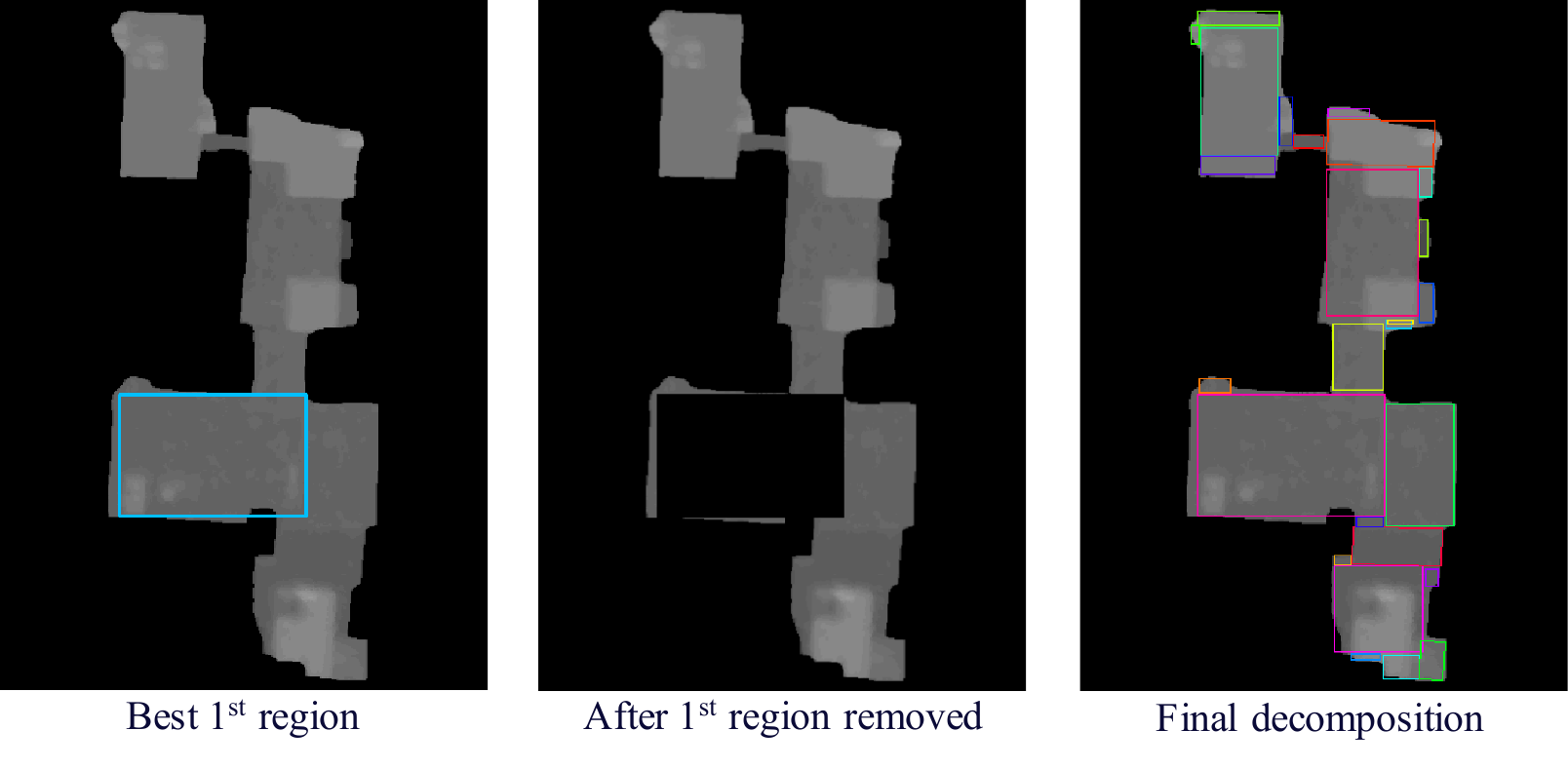}
	\caption{Cascading building decomposition.}
	\label{fig:building_decomposition}
\end{figure}

%\begin{figure}[h]
%	\includegraphics[width=0.5\textwidth]{figure/building_decomposition.pdf}
%	\caption{Building decomposition. Left to right: height map, box decomposition, 3D boxes, original mesh.}
%	\label{fig:building_decomposition2}
%\end{figure}

%
To generate training data for instance segmentation \cite{MaskRCNN}, we use the simulation method mentioned in section \ref{sec:building_simulation}. 
We include both the idealized and noisy boundary images in the training data. 
The simulated buildings are randomly rotated between 0$^\circ$ and 45$^\circ$ so the method is exposed to primitives with arbitrary orientations. 
A total of 10,000 simulated buildings are generated. 
To train the network, a pre-trained model from the COCO dataset is used, and all the layers in the CNN feature extraction are frozen and other layers are trained for 60 epochs. 
Finally, all layers are fine-tuned for another 60 epochs.

To decompose a building into a set of shapes, we cascade the application of the mask R-CNN to partition a building into a set of parts. 
We select one of the instances with the largest intersection over union (IoU) compared with the original mask and then remove that instance from the data. 
This is a greedy approach to decomposing the building into a set of shapes. 
Figure \ref{fig:building_decomposition} demonstrates the decomposition procedure, including the bounding box after the 1st iteration, the image after removing the selected instance, and the final decomposition. 
%
%Figure \ref{fig:building_decomposition2} shows another example, with the height map input, 2D segmented rectangles, 3D boxes, and original mesh.
%
%Note that 3D model has 132 faces, compared to the 63388 faces in the original mesh.
%
%
\subsection{Primitive Fitting}
% 
%%%%%%%%%%%%%%%%%%%%%%%%%%%%%%%%
% Figure: primitive fitting
%%%%%%%%%%%%%%%%%%%%%%%%%%%%%%%%
\begin{figure}[h]
\includegraphics[width=0.5\textwidth]{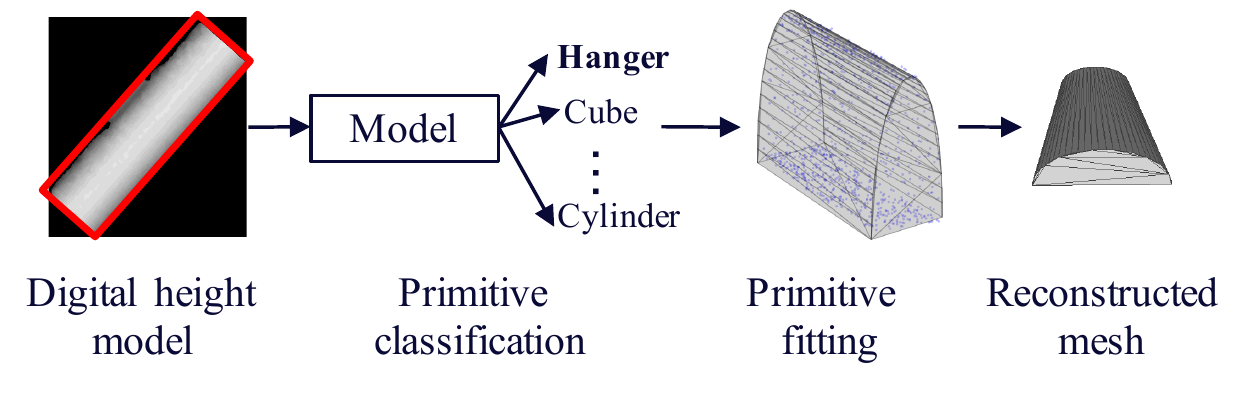}
\caption{Given the bounding box derived from the building decomposition, our primitive fitting pipeline first performs the primitive classification to estimate the roof type, and fit the selected primitive model into the input 3D point cloud.}
\label{fig:primitive_fitting}
\end{figure}
Our primitive fitting pipeline consists of two parts: primitive classification and primitive fitting, where the primitive classification estimates the roof types, and primitive fitting aligns the estimated roof primitive to the input point cloud. The primitive fitting procedure is illustrated in Figure \ref{fig:primitive_fitting}.  
%
%%%%%%%%%%%%%%%%%%%%%%%%%%%%%%%%
% Figure: primitive
%%%%%%%%%%%%%%%%%%%%%%%%%%%%%%%%
\begin{figure}[!h]
	\includegraphics[width=0.5\textwidth]{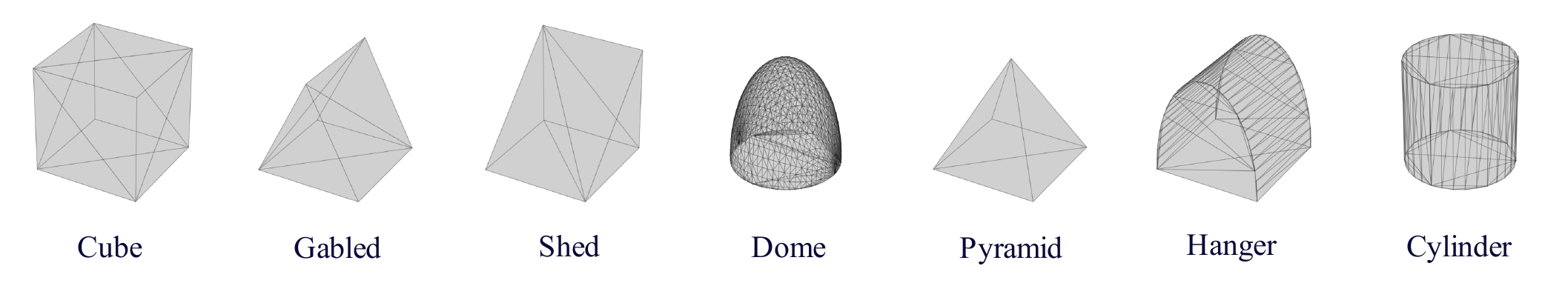}
	\caption{Example roof models in our primitive set.}
	\label{fig:primitive}
\end{figure}
%
%%%%%%%%%%%%%%%%%%%%%%%%%%%%%%%%
% Figure: synthetic point cloud
%%%%%%%%%%%%%%%%%%%%%%%%%%%%%%%%
%\begin{figure}[h]
%	\includegraphics[width=0.5\textwidth]{figure/synthetic_point_cloud.pdf}
%	\caption{We generate synthetic point cloud data by sampling the points on the surface of each roof primitive, and adding random noise on the rotation and height values to %simulate the data distribution of the digital height model.}
%	\label{fig:synthetic_point_cloud}
%\end{figure}
%
\subsubsection{Synthetic point clouds for different roof types}
W use 15 primitive types as our primitive sets, which cover the most common roof types.
For certain roof types, we include different directions, e.g., four directions for shed roof. 
%including shed, gabled, cube, cylinder, cylinder flat, dome, flat, gabled, hanger, parallelogram, rectangular pyramid, shed, square pyramid, triangular pyramid, and mansard.  
%
Example roof primitives are shown in Figure \ref{fig:primitive}.
%
% For certain roof types, we include different directions, e.g., four directions for shed roof. 
%
%For a consistent representation, we fill the missing walls and bottoms for all primitive types. 
%
%Hence, the primitives are watertight.
%
We sample a fixed number (e.g., 2048) of points for each primitive. 
To simulate the digital height model, we add uniform random noise on rotation angle along the z-axis (from -45 to 45 degree) and height values (+/-0.1 in the range of [0, 1]).
%
% We adopt a two-step noise synthesis scheme. We first randomly select a subset of points and add noses on those selected points, which better preserves the point structure and improve the classification performance.
%
We randomly sample 500 point clouds for each primitive; total 7500 synthetic point clouds are used for training and validation.   
\subsubsection{Primitive classification}
We utilize PointNet \cite{PointNet} for primitive classification. 
Given an input point cloud, our method first rectifies the point cloud into the canonical pose, fill the walls and bottoms and normalize the 3D point cloud to an unit cube.
The normalized 3D point cloud is fed into the primitive classification model to estimate the roof type.
The advantages of using the primitive classification are 1) it is more robust to the input point cloud noises and 2) it run faster as avoids fitting each primitive into the point cloud.
\subsubsection{Primitive fitting}
After we obtain the estimated primitive class, we apply Coherent Point Drift (CPD) \cite{CPD} to align the predicted primitive into the target 3D point clouds.
We assume that the transform is rigid, thus the parameter space only involves rotation, translation and scale.
%
%We sample a set of points on the surface of the selected primitive and use these points to fit into the point cloud. 
%
%The optimization is done in an EM-based learning framework. 
%
%In E-step, the algorithm finds which Gaussian the observed point cloud is more likely sampled from, and in M-step it maximizes the negative log-likelihood with respect to transformation parameters. 
%
We compare the fitting results with the initial flat roof model, and select the primitive model with the smallest fitting errors.
\subsection{Texturing}
The last stage of system involves mapping the texture coordinates of the true orthographic color image to the output 3D model.   
The present texturing is limited to the overhead view, and simply wraps the roof texture to the building sides.

\section{Experimental results}
\subsection{Dataset}
\begin{figure}[t]
	\includegraphics[width=0.5\textwidth]{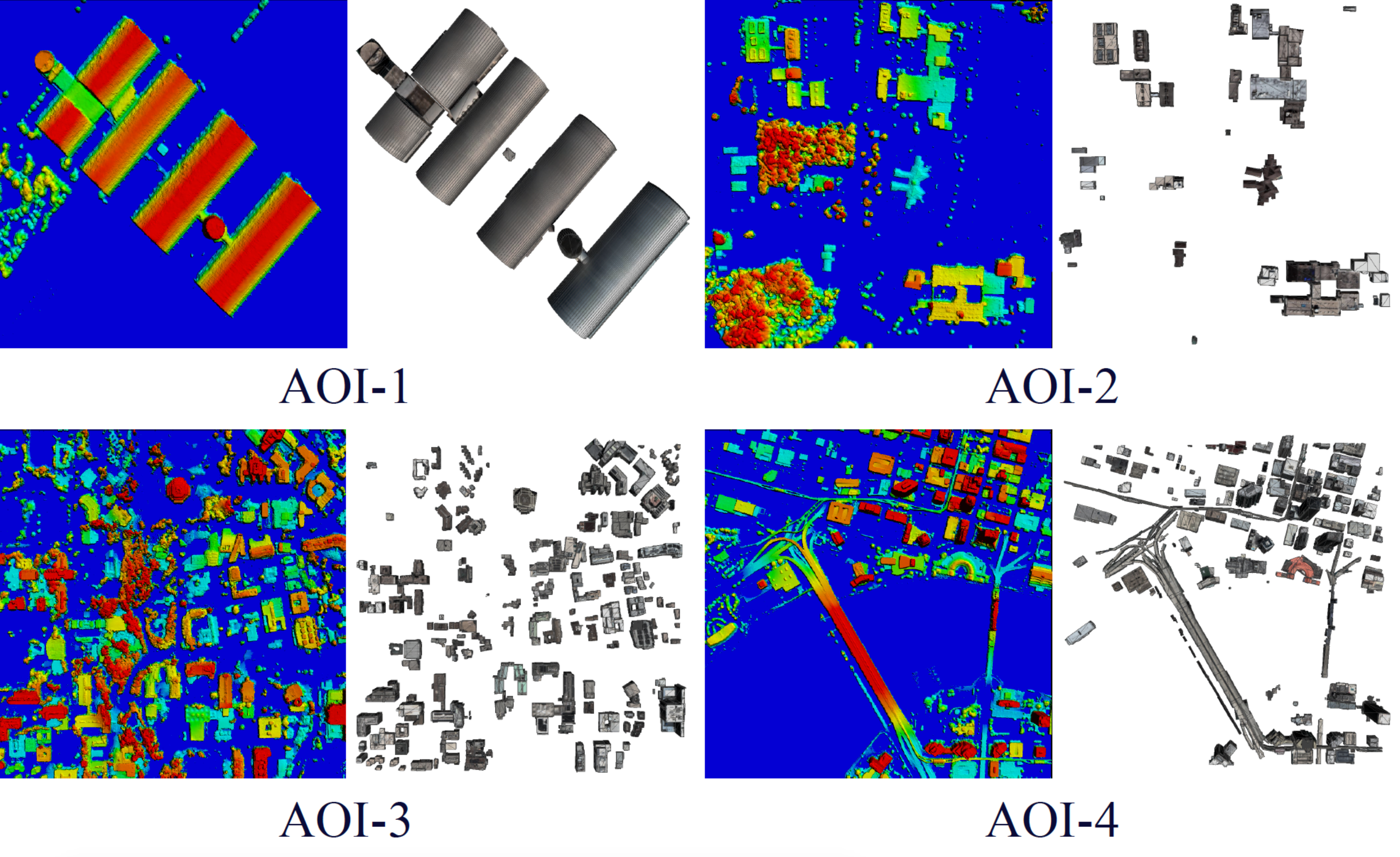}
	\caption{Our building decomposition and primitive fitting results on four testing areas. Left: input digital height model. Right: building decomposition and primitive fitting. Textures are mapped from the true orthographic color images. Zoom-in for better resolution.}
	\label{fig:result}
\end{figure}

We analyze our method on WorldView-3 satellite images. 
The testing areas include four regions: AOI-1 and AOI-2 (Wright-Patterson Air Force Base), AOI-3 (University of California, San Diego), and AOI-4 (Jacksonville, Florida), which have the extent of 0.358, 0.614, 0.962, and 1.813 square kilometers respectively.
We first apply our 2D-based instance building segmentation, 3D reconstruction algorithms to obtain the digital height model for each building region, and 
use it as the input to our primitive-based 3D building modeling approach. 
Figure \ref	{fig:result} shows the reconstructed digital height models for four testing areas.
\subsection{Evaluation Criteria}
We evaluate our method by using the evaluation metrics as described in \cite{MetricEvaluation}. The metrics include completeness, correctness, and Jaccard index in both 2D and 3D space, which are defined as: 
${\rm{completeness}} = {{TP} \mathord{\left/
 {\vphantom {{TP} {(TP + FN)}}} \right.
 \kern-\nulldelimiterspace} {(TP + FN)}}$, 
${\rm{correctness}} = {{TP} \mathord{\left/
 {\vphantom {{TP} {(TP + FP)}}} \right.
 \kern-\nulldelimiterspace} {(TP + FP)}}$, 
 and 
$Jaccard = {{TP} \mathord{\left/
 {\vphantom {{TP} {(TP + FP + FN)}}} \right.
 \kern-\nulldelimiterspace} {(TP + FP + FN)}}$.
We compute the 3D scores by the intersection between our 3D primitive reconstruction and the ground truth digital shape model.
We project the 3D primitive reconstruction into 2D image space, and compute the 2D scores with the ground truth 2D building mask. 
%
%%%%%%%%%%%%%%%%%%%%%%%%%%%%%%%%
% Figure: results
%%%%%%%%%%%%%%%%%%%%%%%%%%%%%%%%
\subsection{Experimental Results}
%%%%%%%%%%%%%%%%%%%%%%%%%%%%%%%%
% Table: experimental results
%%%%%%%%%%%%%%%%%%%%%%%%%%%%%%%%
%
\begin{table}[!h]
\center
\resizebox{0.4\textwidth}{!}{\begin{tabular}{l | |*{4}{c}r}
Metrics			& AOI-1 	& AOI-2 	& AOI-3 	& AOI-4  		\\  \hline
2D completeness      & 93.9 	& 91.5	& 90.7	&  96.0	 	\\ 
2D correctness		& 97.8 	& 91.4	& 89.5      &  88.9            	\\ 
2D Jaccard 		& 92.0 	& 84.3    	& 82.1	& 85.8		\\ \hline
3D completeness  	& 93.7	& 87.5	& 90.0	& 96.0		\\		
3D correctness		& 96.2	& 88.5	& 86.7	& 80.4		\\	
3D Jaccard 		& 90.4	& 78.6	& 79.1	& 77.8		\\  
\end{tabular}}
\caption{Precision, recall and Jaccard index in both 2D and 3D metrics on four testing areas.} 
\label{tb:performance}
\end{table}  

%%%%%%%%%%%%%%%%%%%%%%%%%%%%%%%%
% Table: mesh vertex and face complexity 
%%%%%%%%%%%%%%%%%%%%%%%%%%%%%%%%
\begin{table}[!h]
\resizebox{0.5\textwidth}{!}{\begin{tabular}{l | |*{4}{c}r}
U3D Baseline		& AOI-1 		& AOI-2 		& AOI-3 		& AOI-4  			\\  \hline		
Vertex counts   		& 1,432,900	& 2,458,640	& 3,849,248	& 7,253,844		\\ 
Face counts		& 2,863,407 	& 4,914,139	& 7,694,575    	& 15,076,271    	\\ \\
\bf{Our method}	& AOI-1		& AOI-2   		& AOI-3 		& AOI-4			\\ \hline
Vertex counts 		& 994		& 4822		& 25803		& 18736			\\		
Face counts		& 1680		& 7748		& 38600		& 29072			\\	
\end{tabular}}
\caption{Vertex and triangle face count comparison.} 
\label{tb:face_counts}
\end{table}  

Table \ref{tb:performance} shows the evaluation results on four testing AOIs.
%
%We achieve superior performance compared to other performers;  our method is one of the top two performing methods in phase 1.a IARPA CORE3D challenge.
%
% We beat the phase 1a performance goals, and were one of 2 teams awarded the next phase of the program.
The team will continue working toward improving the performance of the system.
Primitive reconstruction results are shown in Figure \ref{fig:result}.
By representing the polygonal mesh model as a set of primitives, our method significantly reduces the vertex and face numbers of the original dense triangulation to generate a more compact representation (cf. Table \ref{tb:face_counts}).

\section{Discussions}
Based on the fitting results in our four AOIs, we observe the following main challenges.
First, the current building decomposition strategy does not consider the stacked building structures. The next step will be to incorporate stacked structures into the simulation and the decomposition strategy.
%
%Second, currently we assume that each roof model has only one pre-defined slope, we plan to construct a parametric model for each roof type to better model the multiple slopes.
%
Second, there exist some errors in the box decomposition, which will cause problems for primitive fitting. We seek further improvement by considering the constructive solid geometry modeling on our primitive representation.

\section{Acknowledgement}
The research is based upon work supported by the Office of the Director of National Intelligence (ODNI), Intelligence Advanced Research Projects Activity (IARPA), via DOI/IBC Contract Number D17PC00287. The views and conclusions contained herein are those of the authors and should not be interpreted as necessarily representing the official policies or endorsements, either expressed or implied, of the ODNI, IARPA, or the U.S. Government. The U.S. Government is authorized to reproduce and distribute reprints for Governmental purposes notwithstanding any copyright annotation thereon.

\bibliographystyle{IEEEbib}
\bibliography{strings,refs}

\begin{thebibliography}{1}

\bibitem{schnabel2007efficient}
Ruwen Schnabel, Roland Wahl, and Reinhard Klein,
\newblock ``Efficient ransac for point-cloud shape detection,''
\newblock in {\em Computer graphics forum}, 2007.

\bibitem{ren2011minimum}
Zhou Ren, Junsong Yuan, Chunyuan Li, and Wenyu Liu,
\newblock ``Minimum near-convex decomposition for robust shape
  representation,''
\newblock in {\em ICCV}, 2011.

\bibitem{TulsianiCVPR2017}
Shubham Tulsiani, Hao Su, Leonidas~J. Guibas, Alexei~A. Efros, and Jitendra
  Malik,
\newblock ``Learning shape abstractions by assembling volumetric primitives,''
\newblock in {\em CVPR}, 2017.

\bibitem{MaskRCNN}
Kaiming He, Georgia Gkioxari, Piotr Doll\'{a}r, and Ross Girshick,
\newblock ``Mask r-cnn,''
\newblock in {\em ICCV}, 2017.

\bibitem{PointNet}
Charles~Ruizhongtai Qi, Hao Su, Kaichun Mo, and Leonidas~J. Guibas,
\newblock ``Pointnet: Deep learning on point sets for 3d classification and
  segmentation,''
\newblock in {\em CVPR}, 2017.

\bibitem{CPD}
Andriy Myronenko and Xubo Song,
\newblock ``Point set registration: Coherent point drift,''
\newblock {\em TPAMI}, 2010.

\bibitem{MetricEvaluation}
D.~Chilcott H. Goldberg M.~Brown. M.~Bosch, A.~Leichtman,
\newblock ``Metric evaluation pipeline for 3d modeling of urban scenes,''
\newblock in {\em ISPRS Archives}, 2017.

\end{thebibliography}

\end{document}